%% file: root.tex
\documentclass[letterpaper, 10 pt, conference, final]{ieeeconf}  

\IEEEoverridecommandlockouts                              

\usepackage{graphics} 
\usepackage{epsfig} 
\usepackage{mathptmx} 
\usepackage{times} 
\usepackage{amsmath} 
\usepackage{amssymb}  

\usepackage{hyperref}
\usepackage{multirow}
\usepackage{caption}

\title{\LARGE \bf
MC-BEVRO: Multi-Camera Bird Eye View Road Occupancy Detection for Traffic Monitoring}

\author{Arpitsinh Vaghela$^{1}$, Duo Lu$^{2}$, Aayush Atul Verma$^{1}$, Bharatesh Chakravarthi$^{1}$, Hua Wei$^{1}$ and Yezhou Yang$^{1}$%
\thanks{$^{1}$School of Computing and Augmented Intelligence, Arizona State University, Tempe, AZ, USA. Correspondence to \href{mailto:avaghel3@asu.edu}{\tt{avaghel3@asu.edu}}}  
\thanks{$^{2}$Department of Computer Science and Physics, Rider University, Lawrenceville, NJ, USA. {\tt dlu@rider.edu}}
\thanks{For more information, please visit the official MC-BEVRO website: {\tt\href{https://arpitvaghela.github.io/MC-BEVRO/}{\tt{https://arpitvaghela.github.io/MC-BEVRO/}}}}
}%

\begin{document}

\maketitle
\IEEEpeerreviewmaketitle
\thispagestyle{empty}
\pagestyle{empty}

\begin{abstract}
Single camera $3D$ perception for traffic monitoring faces significant challenges due to occlusion and limited field of view. Moreover, fusing information from multiple cameras at the image feature level is difficult because of different view angles. Further, the necessity for practical implementation and compatibility with existing traffic infrastructure compounds these challenges. To address these issues, this paper introduces a novel Bird's-Eye-View road occupancy detection framework that leverages multiple roadside cameras to overcome the aforementioned limitations. To facilitate the framework's development and evaluation, a synthetic dataset featuring diverse scenes and varying camera configurations is generated using the CARLA simulator. A late fusion and three early fusion methods were implemented within the proposed framework, with performance further enhanced by integrating backgrounds. Extensive evaluations were conducted to analyze the impact of multi-camera inputs and varying BEV occupancy map sizes on model performance. Additionally, a real-world data collection pipeline was developed to assess the model’s ability to generalize to real-world environments. The sim-to-real capabilities of the model were evaluated using zero-shot and few-shot fine-tuning, demonstrating its potential for practical application. This research aims to advance perception systems in traffic monitoring, contributing to improved traffic management, operational efficiency, and road safety.

\end{abstract}
\input{introduction}
\input{related_works}
\input{methodology}
\input{experiments}
\input{conclusion}

\bibliographystyle{ieeetr}
\bibliography{root}

\end{document}

%% file: introduction.tex
\section{INTRODUCTION}

The growing number of vehicles has led to many road transportation problems, such as traffic congestion, parking difficulties, environmental impacts, accidents, and safety of pedestrians~\cite{rodrigue2020geography}. Such problems have driven a focus on developing and deploying intelligent transportation systems (ITS). One of the critical technologies in ITS is roadside perception, which converts the raw sensor data to traffic condition information that can help traffic operation decision-making~\cite {xiong2012intelligent}. A core component of such a roadside perception system is $3D$ traffic participant detection, which traditionally uses $3D$ bounding boxes to estimate the object’s $3D$ position, dimension, and orientation in a world reference frame~\cite{arnold2020cooperative}.

However, such object-based traffic participant representation significantly increases the complexity of the detection algorithm. Particularly, it requires a large dataset with manually annotated traffic participants in 3D space, which is time-consuming and labor-intensive to obtain. Also, it is difficult and often unnecessary to identify individual objects in certain cases, such as a cluster of pedestrians or a trailer towed by a truck. On the contrary, it is usually sufficient to know that an area on the road is occupied by such objects. 

Existing works \cite{ye2022rope3d}\cite{yang2023bevheight}\cite{Agrawal2024SemiAutomaticAO} that focus on $3D$ perception from a single view are inherently vulnerable to impairments, including occlusion, restricted perception horizon, and limited coverage. Even if multiple monocular cameras are used in a network, fusing detected objects from different views is error-prone when the individual object detection algorithms running on each view generate conflicting results due to their performance limitation. Additionally, in real-world deployment, traffic monitoring cameras may be mounted at different heights with different pitch angles and coverage, and hence, it is difficult for such object detection algorithms to generalize to different situations.

\begin{figure}[t]
    \centering
    \includegraphics[width=1.0\linewidth]{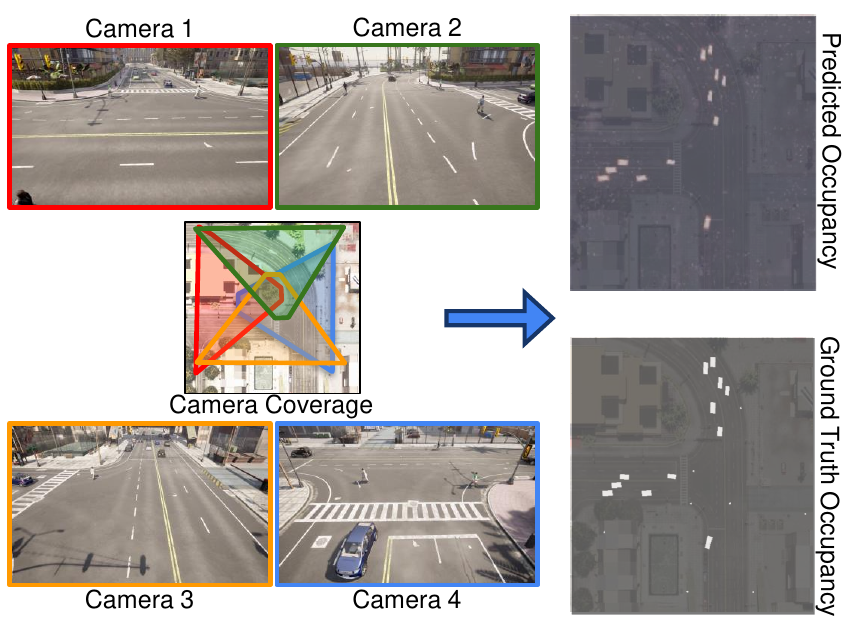}
    \caption{An overview of BEV road occupancy detection using multiple roadside cameras.}
    \label{fig:fig1}
\vspace{-0.25in}
\end{figure}

To address these challenges, we propose a Bird's-Eye-View (BEV) road occupancy detection framework to unify multiple roadside traffic monitoring cameras. In our proposed framework, the image features from each view are transformed into the BEV space and fused together, which increases the overall perception horizon and overcomes occlusion, as shown in Fig. \ref{fig:fig1} (left). Moreover, a traffic scene is represented by an orthographic top view where the region of interest is divided into equidistant cells (i.e., BEV occupancy grid map \cite{bieder2021improving}\cite{schreiber2021dynamic}), with each grid cell encoding a probability on whether it is occupied or not, as shown in Fig. \ref{fig:fig1} (right). While objects on the image may vary in scale due to each camera's image resolution and field-of-view, the BEV occupancy is scale invariant. Moreover, when an object is observed in different views, the fused BEV features can better capture its pose and shape. Additionally, other than road occupancy, such BEV features can be used to predict lane occupancy \cite{kaniarasu2020goal}, traffic flow \cite{kashinath2021review}, and potential collision \cite{yu2018space}. The proposed framework is also compatible with real-world implementation architecture in ITS \cite{arnold2020cooperative}, where multiple cameras are typically linked to a central control box on the roadside, resembling the control system of multiple traffic lights at an intersection.

In summary, we make the following contributions:

\begin{itemize}
    \item We propose a BEV road occupancy detection framework using multiple roadside cameras, implement a late fusion baseline and adapt three early fusion models. We further improve these early fusion models by incorporating backgrounds achieving a 43\% boost in performance.
    
    \item Motivated by the lack of multi-camera BEV occupancy detection dataset, we create a synthetic dataset using the CARLA simulator \cite{dosovitskiy2017carla}. It covers diverse traffic scenes encompassing multiple cameras with varying viewing angles and positions. Our evaluation and ablation studies reveal insights into how the use of multiple cameras and occupancy map resolution impact performance.

    \item To address the practicality and the simulation-to-real-world gap, we develop a pipeline for obtaining real-world data and demonstrate the generalization capabilities of our models using zero-shot and few-shot fine-tuning approaches.
    
\end{itemize}



%% file: related_works.tex
\section{RELATED WORKS}
\label{sec:related_works}
This section reviews existing datasets and methodologies for roadside traffic monitoring followed by multi-camera approaches for BEV people occupancy detection.

\textbf{Roadside Traffic Monitoring Datasets:} Several datasets have been developed to facilitate the study of roadside traffic monitoring. CityFlow \cite{tang2019cityflow} comprises over $3$ hours of synchronized videos of $40$ cameras across $10$ intersections, with $200K$ $2D$ annotations spanning multiple camera views but lacks $3D$ annotations necessary for more advanced tasks like depth estimation and precise localization, limiting its applicability. BoxCars \cite{sochor2018boxcars} offers $116K$ images with $3D$ vehicle annotations from a surveillance camera but is captured from a single view. Rope3D \cite{ye2022rope3d} provides $1.5M$ $3D$ bounding box $50K$ LiDAR scans and images from multiple cameras in a roadside perspective. However, this dataset is limited to non-overlapping individual views, which is unsuitable for studying data fusion from multiple cameras. \cite{yu2022dair} introduces a vehicle infrastructure cooperative dataset consisting of $71K$ LiDAR scans and camera samples with $3D$ annotations. However, since the data is collected for a single traffic scene, it limits the ability to assess the model's generalization to unseen environments with varying camera configurations.

\textbf{3D Perception in Traffic Monitoring:} Detecting traffic participants and localizing them in the 3D space has been a long-term goal for traffic monitoring. Many prior works tried to predict the 3D clues directly from images. BEVHeight~\cite{yang2023bevheight} introduces a robust method for roadside $3D$ object detection by predicting vehicle height instead of depth, which remains consistent across varying distances from the camera. CAROM \cite{lu2023carom} provides a pipeline to detect vehicle $3D$ bounding boxes from segmentation masks and allows object-level fusion from multiple monocular cameras. In \cite{arnold2020cooperative} different fusion techniques like early, late, and hybrid are explored to optimize $3D$ perception using both camera and LiDAR data. TransIFF~\cite{chen2023transiff} leverages a transformer-based fusion framework to address the domain gap and enhance feature fusion robustness. Similarly, the Feature Flow Net~\cite{yu2024flow} uses a flow-based feature fusion approach for cooperative $3D$ object detection. Additionally, MonoUNI~\cite{jinrang2024monouni} introduces normalized depth as a unified optimization target, leading to top performance in infrastructure-side benchmarks without relying on additional data inputs. However, these methods focus on object-level detection, which is limited by its focus on individual objects, struggles with occlusions, and lacks the ability to handle rare and irregular objects, such as trailers towed by a truck or a crowd of pedestrians. In contrast, BEV occupancy detection offers a more comprehensive spatial representation, handling occlusions and crowded environments more effectively and providing better generalization across diverse scenarios.


\textbf{BEV People Detection from Multiple Cameras:} The concept of predicting occupancy probabilities from multiple views using a discrete occupancy map was first introduced for pedestrian detection \cite{fleuret2007multicamera}. The occupancy map corresponds to a discrete-state (binary) random field. Given a $3D$ world space observed by the cameras, the BEV occupancy map divides the $XY$ plane of $3D$ space into discrete vertical columns of resolution. The presence or absence of objects within this $3D$ space is then represented on the map using binary encoding. The resolution parameters, $\Delta x\, \text{ and }\, \Delta y$, dictate the level of detail and granularity of the occupancy information captured in the map, enabling a balance between precision and computational efficiency in processing and analysis. \cite{fleuret2007multicamera}\cite{baque2017deep} combine the convolutional neural net (CNN) network with conditional random fields (CRF) to predict BEV occupancy. Deep-MCD \cite{chavdarova2017deep} uses CNN architecture followed by multilayer perceptron over the generated feature map to predict the BEV occupancy. MVDet \cite{hou2020multiview} proposes an anchor-free approach to aggregate multiview. It first applies a perspective transformation, followed by concatenating the projected feature maps from multiple viewpoints and performing large kernel convolution for spatial aggregation. MVDeTr \cite{hou2021multiview} introduces the concept of a shadow transformer, which uses a deformable transformer on multiple camera images for spatial aggregation. SHOT \cite {song2021stacked} introduced a combination of homographies at different heights to improve the quality of projections. GMVD \cite{vora2023bringing} focuses on generalization ability across multiple camera configurations by using average pooling. These methods have been applied to people detection in open spaces, where each person is typically assumed to occupy a single cell. In contrast, detecting traffic participants in traffic monitoring requires handling multi-cell occupancy due to larger size of vehicles, adding complexity to the spatial representation and detection process.

%% file: methodology.tex
\section{Multi-camera BEV Occupancy Dataset}
\label{sec:datagen}

\begin{figure}[t]
    \centering
    \includegraphics[width=1\linewidth]{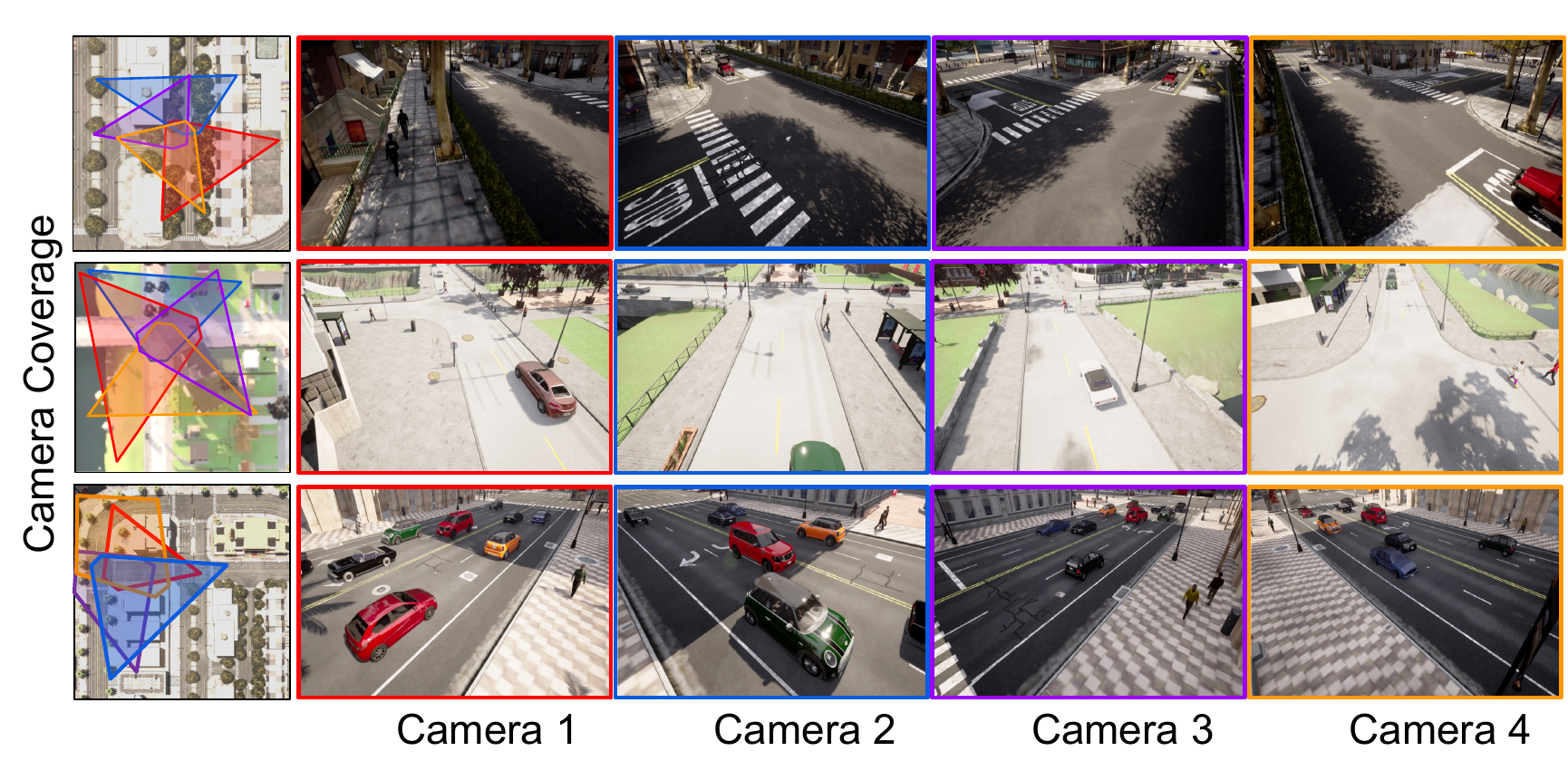}
    \caption{Samples from the generated multi-camera occupancy dataset showcasing diverse scenes and perspectives.}
    \label{fig:fig2}
\vspace{-0.4in}
\end{figure}

To the best of our knowledge, there is no existing open dataset for multi-camera road occupancy detection for traffic monitoring on the roadside or from the road infrastructure. In this paper, we utilized CARLA~\cite{dosovitskiy2017carla} to generate a synthetic dataset. Town, 10 of the simulator, was chosen as the primary focus due to its larger urban area, numerous intersections, wide varieties of traffic scenes in an urban environment, flexible locations for pedestrians and vehicle generation, and close resemblance of real-world traffic patterns. Twelve traffic scenes were used from this town with four cameras at each scene. Additionally, we also included six other towns in the simulator for data collection, which further improves the diversity in traffic scenarios. Some of the six towns had distinct visual features, such as light-colored asphalt, which can negatively affect model performance on real-world data with dark road surfaces. Eight scenes were used from these six towns with the same camera settings. For each scene, we consulted road transportation experts from the Institute of Automated Mobility \footnote{\url{https://www.azcommerce.com/iam/}} and strategically placed the four cameras to monitor road traffic (mainly for vehicles) within a region of interest. Specifically, the cameras were positioned at a height varying from $5\,\text{m to }\,8\,\text{m}$, covering a distance roughly from $50\,\text{m} - 150\,\text{m}$. Each camera had a resolution of $1920\times1080$ and a field of view of $90^{\circ}$ capturing data at $20\,\text{ frames per second}$. The degree of camera overlap varies across the scenes to evaluate the system's ability to aggregate information from multiple views and ensure consistent detection. Examples are shown in Fig. \ref{fig:fig2}.


For each simulated vehicle and pedestrian in CARLA, its position $(x, y)$, dimension $(l, w)$, orientation $(\theta)$, and the $2D$ bounding boxes on each camera view are stored.
To generate the ground-truth BEV road occupancy at a traffic scene, we first specify a reference origin $(x_0,y_0)$ and determine the total area in the BEV space covered by all four cameras. Next, we use a predefined occupancy grid size (e.g., $480\times480$) and the cell's metric resolution $(\Delta x, \Delta y)$ to define the BEV grid boundary based on the actual camera coverage. This step is critical for maintaining consistent cell resolution in meters across the dataset so that our models can be generalized to a new traffic scene. Finally, we use the calculated cell resolution and map size to transform each vehicle's position $(x, y)$ and dimension $(l, w)$ from the world reference frame to our BEV space position $(g_x, g_y)$ and dimensions $(g_l, g_w)$ as shown in the following equations. Finally, we generate ground-truth binary occupancy by rasterizing the objects in the BEV grid space. An example is shown in Fig. \ref{fig:fig1}.

\begin{align}
\label{eq:grid_generation}
    (g_x, g_y) &= (\frac{x - x_0}{\Delta x}, \frac{y - y_0}{\Delta y}) &
    (g_l, g_w) &= (\frac{l}{\Delta x}, \frac{w}{\Delta y})    
\end{align}

This procedure is conducted for each image, and the generated data consists of $500$ images from each camera at each scene. Our dataset has 8 three-way intersections, 6 four-way intersections, and 6 road segments, which ensures a balanced representation of urban environments with different traffic dynamics. In total, the dataset has $160K$ traffic participants with $70K$ vehicle instances and $90K$ pedestrian instances.  

\section{Multi-Camera BEV Occupancy Detection}
\label{sec:baseline_methodologies}

This section details the proposed framework for BEV occupancy detection using multi-camera inputs for traffic monitoring. We first describe the framework's inputs and outputs, followed by an explanation of the late fusion approach and three early fusion methods. Moreover, we demonstrate that incorporating static background images from the cameras can further enhance model performance and generalization across different scenes.

\subsection{Input and Output}

The proposed framework takes images from four static cameras and their calibration as input. These images are processed to generate two BEV occupancy grid maps: one for vehicles and the other for pedestrians. Each occupancy grid map represents a BEV representation of the scene, where the area is divided into discrete cells, and each cell in the map indicates a probability of whether it is occupied. Pedestrians are typically represented by Gaussian centered at a single cell, while vehicles are represented by multiple cells due to their larger size. This occupancy representation provides a comprehensive view of the environment without focusing on individual object detections.

\begin{figure}[t]
    \centering
    \includegraphics[width=0.98\linewidth]{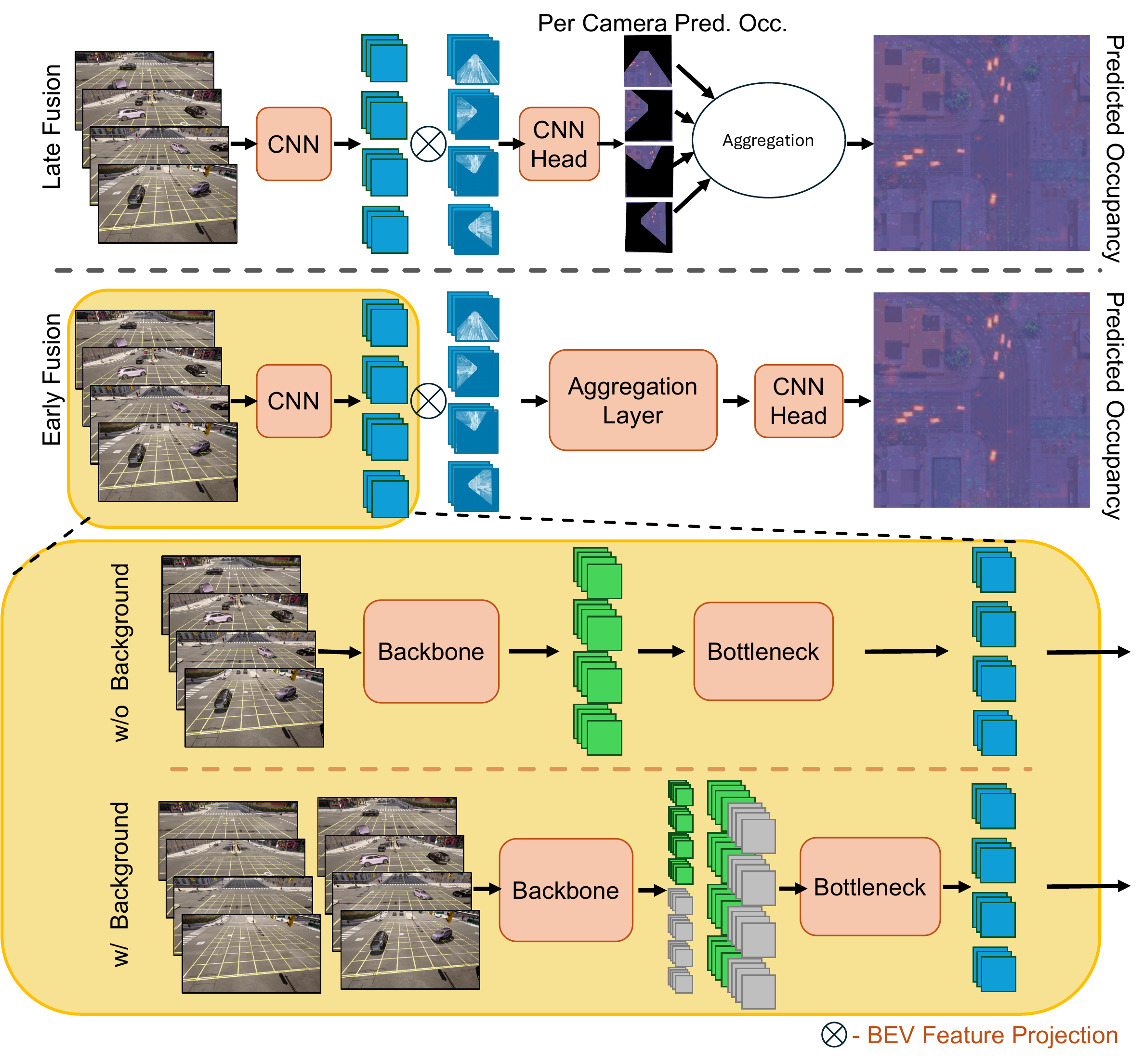}
    \caption{An overview of implemented models (top) the baseline late fusion model predicts per-camera occupancy maps, fused via mean aggregation, (middle) the early fusion model fuses projected features using a neural network, (bottom) the integration of background information is depicted.}
    \label{fig:fig3}
\vspace{-0.25in}
\end{figure}
\subsection{Late Fusion Approach}

A late fusion approach is implemented where images from each camera are processed independently. As shown in \figureautorefname~\ref{fig:fig3}~(top), each image passes through a ResNet-18~\cite{resnet} backbone, where the last three stridden convolutions are replaced by dilated convolutions to enhance spatial resolution. This step extracts features from each view, which are then transformed using a homography projection onto the ground plane of the BEV space. A final convolution layer predicts the BEV occupancy for the area covered by each camera, which is then fused using mean aggregation. The model shares weights across all camera views to ensure consistency, and the independent predictions are combined to generate the final occupancy map.

\subsection{Early Fusion Approaches}

In addition to the late fusion approach, we developed three early fusion methods with a similar architecture as shown in \figureautorefname~\ref{fig:fig3}~(middle). All three methods utilize the same ResNet-18 backbone and generate the two occupancy maps in the same format. However, a bottleneck layer is used to reduce the total number of feature maps, and each early fusion method employs a unique feature aggregation technique in the BEV space, shown as the aggregation layer in the figure. They are adapted from prior BEV human detection works as follows.

\textit{MVDet} \cite{hou2020multiview} utilizes three layers of dilated fully convolutional layers on the projected feature maps to use a relatively large receptive field to consider the ground plane neighbors jointly. We adapt these convolutional layers as the aggregation layer.

\textit{MVDeTr} \cite{hou2021multiview} uses a deformable transformer on a projected feature map obtained from multi-camera images, where each point in the resultant feature map is obtained by applying an attention mechanism on a fixed set of reference points across the projected feature maps of all views for each attention head. We adapted this transformer in our BEV space as the aggregation layer.

\textit{GMVD} \cite{vora2023bringing} uses average pooling for spatial aggregation of the projected feature maps. We adapted the same average pooling method as the aggregation layer.

\subsection{Improved Generalization via Background Integration}

To enhance generalization across different scenes, we integrate static background images from each camera view into the framework as illustrated in \figureautorefname~\ref{fig:fig3}~(bottom). These static background images are passed through the same ResNet-18 backbone used for processing the current images. In simulation, they are just static images when no traffic participants are spawned. In practice, they can be computed through an online rolling background subtraction method. The backbone generates a feature map of shape $(C, H, W)$ for each of the $N$ current images and their corresponding $N$ background images, resulting in a combined $2N$ feature map of shape $(2N, C, H, W)$. The features from the current and background images are then concatenated along the channel dimension, transforming the feature map from $(2N, C, H, W)$ to $(N, 2C, H, W)$. This concatenated map is subsequently passed through a bottleneck layer consisting of a $1\times1$ convolution, which reduces the feature dimensions to $(N, C’, H, W)$. Each output channel from the bottleneck layer is a linear combination of the $2C$ channels corresponding to the current image and background features \cite{lin2013network}. The resulting features are then fed into the aggregation layers of the early fusion methods, producing two BEV occupancy maps for vehicles and pedestrians. Additionally, background features can be cached and reused to reduce computational overhead during inference, which can significantly speed up the process.

%% file: experiments.tex
\begin{figure*}[t]
    \centering
    \includegraphics[width=0.92\textwidth]{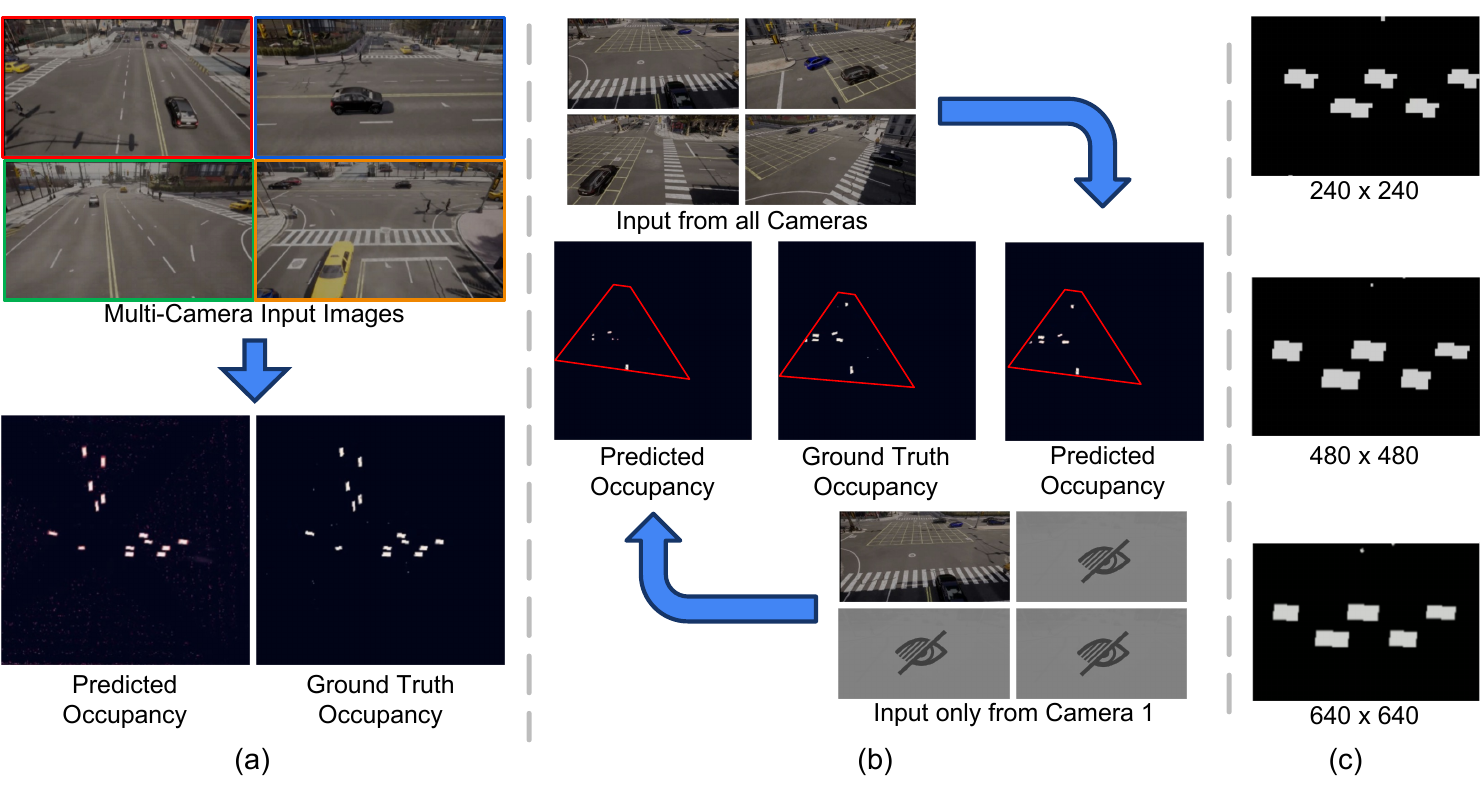}
    \caption{Qualitative results of experiments comparing (a) predicted occupancy against ground truth for corresponding multi-camera input, (b) predicted occupancy for the area covered by one camera, comparing input from one camera versus input from all cameras, and (c) better resolution in occupancy map with increasing grid size.}
    \label{fig:fig4}
\vspace{-0.2in}
\end{figure*}


\section{EXPERIMENTAL EVALUATION}
\label{experiments}

\subsection{Experiment Setup and Model Training}

We establish a late fusion baseline and compare it with adapted early fusion methods: MVDeTr, MVDet, GMVD. We also augment each early fusion method with the background (denoted as ``+BG''). These models were trained on $16$ scenes of the generated dataset and evaluated on four unseen scenes, two from Town 10 and two from other towns. Each model was trained on one single NVIDIA A100 GPU.

The occupancy map size was set to 480$\times$480 by default, with a resolution of $0.31\,\text{m}$ $\times$ $0.31\,\text{m}$ per grid cell. All early fusion models were trained with a batch size of $2$ for $8$ epochs at a learning rate of $0.005$. The late fusion baseline is trained at a higher learning rate of $0.05$ and a batch size of $1$, which generated the best result in hyperparameter tuning.

\subsection{Evaluation Metric}
\vspace{-0.02in}
For all experiments, intersection over union (IoU), also known as the Jaccard index~\cite{costa2021further}, is used to quantitatively measure the model's performance by calculating the overlap between the predictions and the ground truth occupancy. 
The predicted probabilities are threshold at $0.5$ to obtain the predicted binary occupancy map $M$, which is compared with the ground truth binary occupancy map $O$ to obtain the IoU,

\begin{equation}
    \text{IoU} = \frac{\sum_{x,y}MO}{\sum_{x,y}M + \sum_{x,y}O - \sum_{x,y}MO}.
\end{equation}

\subsection{Performance Evaluation and Model Comparison}

As shown in \tableautorefname~\ref{tab:model_results}, the late fusion model, serving as the baseline, performs the worst with an IoU of 0.20388, indicating its limitations in capturing the necessary spatial relationships. In comparison, all early fusion models significantly outperform the baseline, demonstrating the effectiveness of early fusion techniques. Particularly, the simple average pooling method adapted from GMVD has marginally better results than MVDet and MVDeTr. Furthermore, adding background information to these models consistently boosts the performance. Here, (MVDeTr + BG) achieved the best IoU of 0.67759, which is a 43\% improvement, followed by (MVDet + BG) and (GMVD + BG).  This highlights the crucial role of background information in generalization to new unseen scenes. An example qualitative result is shown in Fig. \ref{fig:fig4} (a) with (MVDeTr + BG).

\begin{table}[h]
\centering
\begin{tabular}{rccc}
\hline
\textbf{Model} & \textbf{IoU (Model)} & \textbf{IoU (Model + BG)} \\ \hline
Early Fusion (GMVD) & \textbf{0.49454} & 0.60671 \\
Early Fusion (MVDet) & 0.48755 & 0.66986 \\ 
Early Fusion (MVDeTr) & 0.473734 & \textbf{0.67759} \\
\hline
Late Fusion & 0.20388 & -\\

\hline
\end{tabular}
\caption{Performance results of proposed models}
\label{tab:model_results}
\vspace{-0.2in}
\end{table}

\begin{figure*}[!th]
    \centering
    \includegraphics[width=0.94\linewidth]{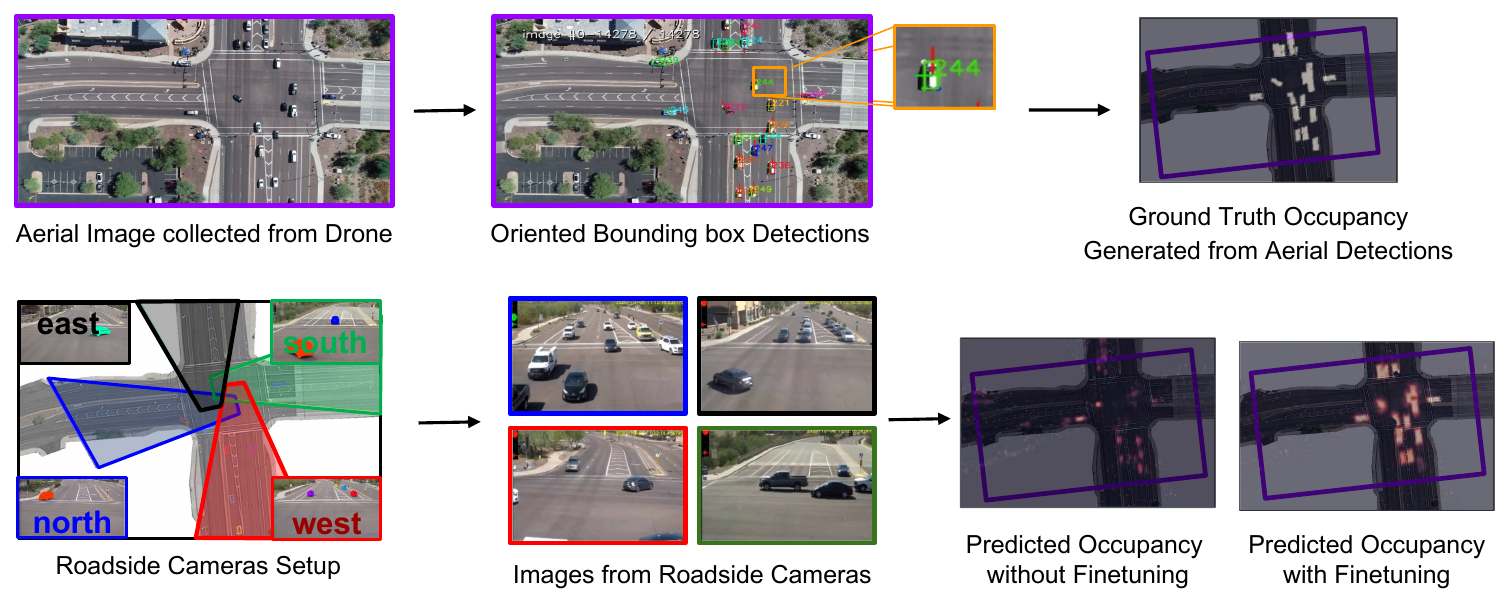}
    \caption{Real-world data collection. The top row shows the pipeline to get BEV occupancy from aerial images using oriented bounding box detections. The bottom row presents camera setup and predicted BEV occupancy with and without fine-tuning.}
    \label{fig:fig5}
\vspace{-0.2in}
\end{figure*}
\subsection{Ablation Study: Impact of Multi-camera Input}

The impact of multi-camera input on the performance was assessed by evaluating the trained models over the area covered by a single camera with input from that camera against input from all four cameras, as shown in Fig.~\ref{fig:fig4} (b). The results are shown in \tableautorefname~\ref{tab:view_results}. The results clearly show that using images from multiple cameras can help the models predict the occupancy map better, which validates the need for multi-camera input for better information fusion.

\begin{table}[h]
\centering
\begin{tabular}{rcc} \hline
\textbf{Model}  & \textbf{Camera 1 Only }   & \textbf{All Cameras} \\ \hline
GMVD        & 0.4146                       & 0.4984         \\ 
GMVD + BG   & 0.1739                       & 0.6076         \\ 
MVDet       & 0.2644                        & 0.4932          \\
MVDet + BG  & 0.3139                       & 0.7319          \\
MVDeTr      & 0.2438                      & 0.5031          \\
MVDeTr + BG & \textbf{0.5047}             & \textbf{0.7544}          \\
\hline
\end{tabular}
\caption{Performance comparison of single-camera vs. multi-camera. IOU results are only computed under the area covered by camera 1 for fair comparison. See Fig.~\ref{fig:fig4} (b) for an example.}
\label{tab:view_results}
\vspace{-0.1in}
\end{table}

\subsection{Ablation Study: Impact of Different Occupancy Map Sizes}

We further explored the impact of varying BEV occupancy map resolution and size on the model's performance by training additional early fusion models with (MVDeTr + BG). Examples of the effects of the different occupancy map sizes on ground truth occupancy is shown in Fig. \ref{fig:fig4} (c). The results are shown in \tableautorefname~\ref{tab:grid_results}.

\begin{table}[h!]

\centering
\begin{tabular}{c|c|c|c|c}
\hline
\textbf{Occ. map size} & 240$\times$240 & 320$\times$320 & 480$\times$480 & 640$\times$640 \\ \hline
\textbf{MVDeTr + BG}      & 0.5480  & 0.5659  & \textbf{0.6776}  & 0.6220  \\ \hline
\end{tabular}
\caption{Impact of the BEV grid resolution.}
\label{tab:grid_results}
\vspace{-0.2in}
\end{table}

From the table, the performance peaks at the grid size of 480$\times$480. On the one hand, we believe that the rasterization error limits the performance at low BEV grid resolution, as illustrated in Fig.~\ref{fig:fig4} (c). On the other hand, we speculate that the model struggles with precision but has good recall at high BEV grid resolution. One reason is that vehicles are usually not boxy around the corners, and another reason may be the fact that the chassis of vehicles are lifted above the ground by the suspension system, which makes it difficult for the neural network to precisely find the exact boundaries of the 3D bounding boxes projected on the ground from the image features. Additionally, the gains in performance with higher grid resolution come at the cost of increased computational requirements, which is an important consideration for deploying such models in real-world scenarios.

\subsection{Real-World Evaluation}

We assessed the ability of our models trained on synthetic data to generalize to a real-world scene. With the assistance of the Maricopa County Department of Transportation in Arizona, we obtained synchronized videos from a system of four infrastructure-based traffic monitoring cameras \cite{lu2021carom} deployed at an intersection. We also flew a drone and developed a vehicle localization pipeline \cite{lu2023carom} from aerial videos, as shown in Fig.\ref{fig:fig5}.

We selected three and a half minutes of video taken at $30$ fps, collected simultaneously from the four infrastructure cameras and the drone to evaluate the real-world performance of our proposed framework. This data was then time-synchronized, yielding $6400$ data samples, where each data sample contains four synchronized images and the reference measurement from the drone. The occupancy grid map size and image size were kept consistent with our synthetic dataset, but the grid cell resolution was reduced to $0.5\,\text{m}\times 0.5\,\text{m}$ because the infrastructure-based cameras cover a larger area. Our drone-based method has a vehicle location accuracy of $0.1\,\text{m}-0.3\,\text{m}$ \cite{lu2023carom}, which is approximately half the size of one cell in our BEV grid space. Hence, it is adequate to be used as the ground truth.

The early fusion methods with background information were evaluated in both zero-shot and few-shot fine-tuning settings with 24 samples. Two fine-tuning strategies were designed to test different aspects of the model's performance. In the first setting, the backbone and bottleneck were frozen while the aggregation layer and the head were fine-tuned. This approach is intuitive but it does not address the sim-to-real gap in the backbone features. In the second setting, the aggregation layer and head were frozen while the backbone and bottleneck were fine-tuned. This is the typical case in practice and it is designed to study the domain adaptation capability as well as the limitation of the aggregation layers.

\begin{table}[h!]
\centering
\resizebox{\columnwidth}{!}{%
\begin{tabular}{rccc}
\hline
\textbf{Model} &
  \textbf{Zero Shot} &
  \textbf{\begin{tabular}[c]{@{}c@{}}Few Shot \\ (Frozen Backbone)\end{tabular}} &
  \textbf{\begin{tabular}[c]{@{}c@{}}Few Shot \\ (Frozen Head \& Agg.)\end{tabular}} \\ \hline
GMVD + BG   & 0.0282  & 0.28408 & \textbf{0.67013} \\
MVDet + BG  & 0.1180  & \textbf{0.64481} & 0.64348 \\
MVDeTr + BG & \textbf{0.19141} & 0.64316 & 0.66936 \\
\hline
\end{tabular}%
}
\caption{Performance evaluation of Models in real-world under zero-shot and few-shot fine-tuning settings.}
\label{tab:few-shot-table}
\vspace{-0.2in}
\end{table}

The results indicate that all models show limited zero-shot generalization performance to real-world data, while both few-shot fine-tuning strategies improve performance significantly with only 24 data samples. Particularly, when fine-tuning the backbone and bottleneck while freezing the aggregation layer and head, the simple method with average pooling in the aggregation layer (GMVD) achieves marginally better results than the other two methods with lots of learnable parameters in the BEV space, placing greater reliance on the backbone for domain adaptation. In contrast, MVDeTr and MVDet distribute this burden between the backbone and the aggregation layer. These findings highlight the models' capability to generalize from synthetic to real-world data with fine-tuning. We speculate that it might be sufficient to fine-tune the backbone in other simpler tasks to address the sim-to-real gap, such as semantic segmentation in the image domain with a single camera, and this will be explored in our future work.




%% file: conclusion.tex
\section{CONCLUSION}
\label{conclusion}

This paper introduces a novel framework for multi-camera Bird-Eye-View (BEV) road occupancy detection using roadside cameras, addressing challenges such as occlusion and limited field of view in traffic monitoring. We developed and implemented three different early fusion models, with proposed background integration to further boost the performance. To address lack of dataset, we created a synthetic dataset using CARLA, featuring diverse scenes and camera setups. An evaluation with real-world data is also provided, demonstrating our model's ability to generalize to real-world scenes, with few shot fine-tuning further improving performance. Our contributions advance perception in traffic monitoring, provide valuable research data, and offer a practical approach for downstream traffic analysis applications.
